\documentclass{ecai} 

\usepackage{latexsym}
\usepackage{amssymb}
\usepackage{amsmath}
\usepackage{amsthm}
\usepackage{booktabs}
\usepackage{enumitem}
\usepackage{graphicx}
\usepackage{algorithm}
\usepackage{algorithmic}
\usepackage{times}
\usepackage{latexsym}
\usepackage{makecell}
\usepackage{graphicx}
\usepackage{amsmath,amsthm}
\usepackage{amssymb}
\usepackage{xcolor}
\usepackage{soul}
\usepackage{url}
\usepackage{hyperref}
\usepackage{soul,xcolor}
\usepackage{siunitx}
\usepackage{booktabs}
\usepackage{multirow}
\usepackage{graphicx}
\usepackage{subfig}
\usepackage{color}
\usepackage{url}
\usepackage{hyperref}

\newcommand{\BibTeX}{B\kern-.05em{\sc i\kern-.025em b}\kern-.08em\TeX}

\begin{document}


\begin{frontmatter}


\title{{\em Two eyes, Two views, and finally, One summary!} Towards Multi-modal Multi-tasking Knowledge-Infused Medical Dialogue Summarization}


\author[A]{\fnms{Anisha}~\snm{Saha}\footnote{Corresponding Author. Email:anisha0325@gmail.com}}
\author[A,B]{\fnms{Abhisek}~\snm{Tiwari}}
\author[B]{\fnms{Sai}~\snm{Ruthvik}}
\author[A]{\fnms{Sriparna}~\snm{Saha}}

\address[A]{Indian Institute of Technology, Patna}
\address[B]{Clinical AI Assistance}

\begin{abstract}
    We often summarize a multi-party conversation in two stages: chunking with homogeneous units and summarizing the chunks. Thus, we hypothesize that there exists a correlation between homogeneous speaker chunking and overall summarization tasks. In this work, we investigate the effectiveness of a multi-faceted approach that simultaneously produces summaries of medical concerns, doctor impressions, and an overall view. We introduce a multi-modal, multi-tasking, knowledge-infused medical dialogue summary generation ({\em MMK-Summation}) model, which is incorporated with adapter-based fine-tuning through a gated mechanism for multi-modal information integration. The model, {\em MMK-Summation}, takes dialogues as input, extracts pertinent external knowledge based on the context, integrates the knowledge and visual cues from the dialogues into the textual content, and ultimately generates concise summaries encompassing medical concerns, doctor impressions, and a comprehensive overview. The introduced model surpasses multiple baselines and traditional summarization models across all evaluation metrics (including human evaluation), which firmly demonstrates the efficacy of the knowledge-guided multi-tasking, multimodal medical conversation summarization. The code and dataset are available at \url{https://github.com/NLP-RL/MMK-Summation}.
\end{abstract}
\end{frontmatter}

\section{Overview}
\begin{figure}[t]
    \centering
    \includegraphics[width=\columnwidth]{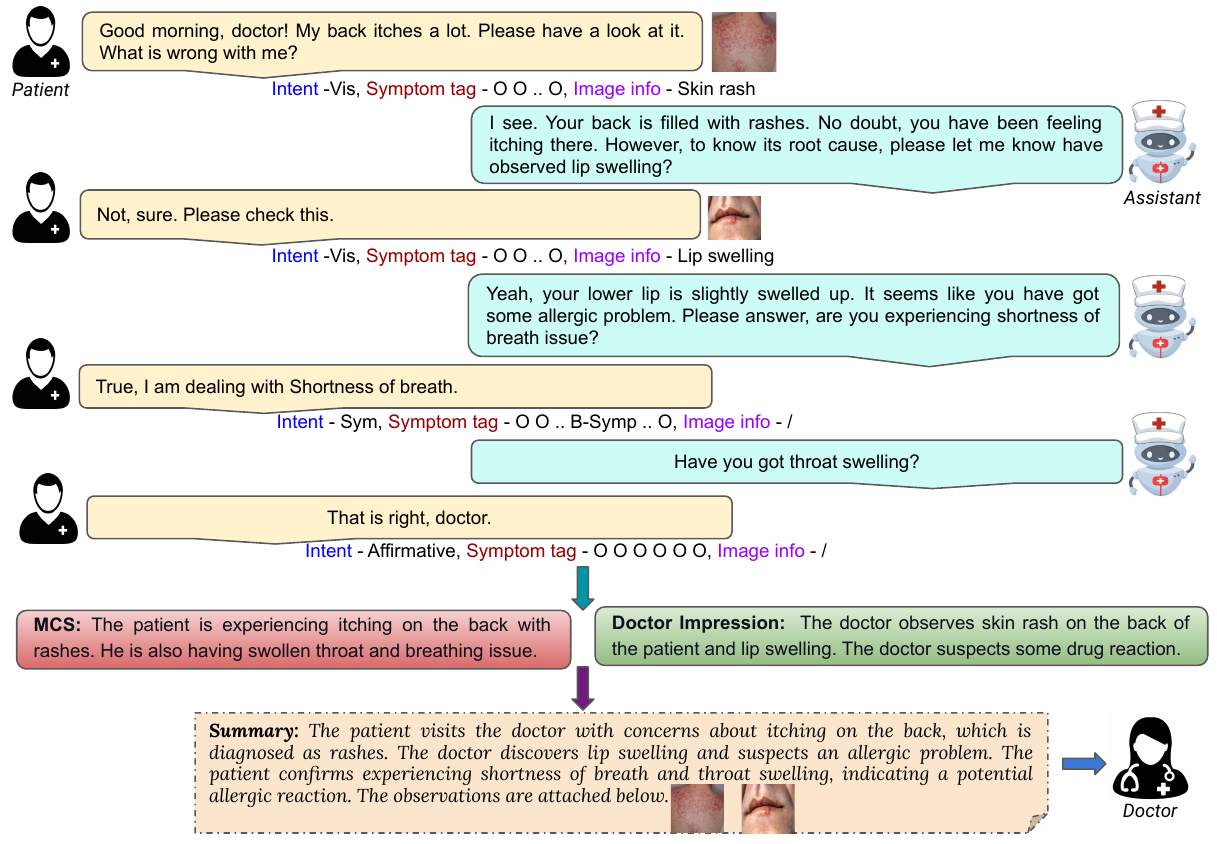}
    \caption{Importance of clinical conversation summarization and the two-view approach for producing an effective summary}
    \label{F1_Problem}
    \vspace{2em}
\end{figure}

Healthcare services are a fundamental necessity for individuals regardless of their location, social background, or beliefs. Seeking medical guidance is our typical response to health issues. Nevertheless, the state of healthcare, especially in government hospitals, is currently distressing in terms of patient management. In numerous developing nations, waiting several days for a doctor's consultation has become the norm due to an inadequate number of healthcare staff. A recent report from the World Health Organization anticipates a shortage of 30 million healthcare workers by 2030 relative to the growing population\footnote{\url{https://www.who.int/health-topics/health-workforce}}. Moreover, a significant proportion of the population, surpassing one-third, resides in rural regions\footnote{\url{https://www.statista.com/statistics/1328171/rural-population-worldwide/}} where accessing medical professionals remains challenging. To address these challenges, the utilization of Artificial Intelligence (AI) based tools and techniques has surged dramatically in the past few years \cite{chang2019integration, lin2021graph}. One such manifestation is automated symptom investigation and diagnosis (Figure \ref{F1_Problem}) to assist doctors in conducting preliminary assessments.

Within a digital healthcare framework, a patient interacts with various stakeholders, each seeking distinct aspects of the doctor-patient dialogue: some require the patient's primary concern, while others seek the doctor's perspective or an overarching summary. In this regard, we bring forth a few concepts, starting with Medical Concern Summary (MCS), which aims to create a concise overview of the patient's key issue. Doctor Impression (DI) encapsulates the concluding reaction and impression of the doctor subsequent to the conversation with the patient. In the process of clinical diagnosis and treatment, a patient's journey typically involves multiple interactions rather than a single visit. Consequently, reviewing the entire transcript of a previous lengthy conversation can be time-consuming. Therefore, having access to the patient's medical concern summary (MCS) along with the doctor's impression (DI) serves as a helpful synopsis/action point of the case for different healthcare stakeholders, reducing the need to refer to the lengthy transcript. An example of both MCS and DI are illustrated in Figure \ref{F1_Problem}. Motivated by the efficacy and importance of MCS and DI, we attempt to study their impact on overall summary generation and model this information as a crucial insight while generating an overall summary of discussion between clinicians and patients.

In our daily lives, the act of summarizing documents is a common task, and the method employed can vary depending on the context. Nonetheless, a general approach that many of us tend to follow does exist. As humans, our strategy for summarizing a dyadic conversation diverges from a strictly sequential process. We initiate this process by grasping the broader context, after which we assimilate pertinent specifics from diverse utterances. This enables us to craft a succinct summary that remains coherent with the overarching context. Following this initial overview, we take into account the perspectives of each speaker, which leads us to formulate a comprehensive summary that effectively captures the central elements. For example, when summarizing the dialogue (Figure \ref{F1_Problem}), the initial step involves reading the entire conversation. Subsequently, we identify the central patient concern, referred to as the medical concern summary (MCS), along with the doctor's response, termed the doctor's impression (DI). Ultimately, the summary of the dialogue is crafted, focusing on essential elements (MCS and DI), resulting in a coherent and effective summary. Motivated by the observation, we aim to investigate some fundamental research questions related to multi-modal medical dialogue summarization and propose a two-phase, knowledge-infused multi-modal medical dialogue summarization framework.

\hspace{-0.33cm}\textbf{Research Questions} In this paper, we aim to investigate the following three research questions: RQ1: \textit{Is there any correlation between medical concern summary (MCS) generation and overall dialogue summarization?}, RQ2: \textit{Is there any correlation between doctor impression (DI) generation and overall dialogue summarization?}, RQ3: \textit{Are these three tasks correlated?  Would learning all three simultaneously result in an improved overall summary generation?}

\begin{table*}[hbt!]
    \centering
        \caption{Performances of different baselines and proposed models for multi-modal clinical conversation summary generation}
    \scalebox{0.97}{
    \begin{tabular}{lccccccccccc}
    \hline
     \textbf{Model}  & \textbf{B-1} & \textbf{B-2} & \textbf{B-3} &\textbf{B-4} & \textbf{BLEU}  & \textbf{R- 1}  & \textbf{R- 2} & \textbf{ROUGE- L}  & \textbf{METEOR} & \textbf{Jaccard Sim} & \textbf{BERT Score} \\ \hline
     GPT-2 \cite{budzianowski2019hello}  & 11.65 & 5.34 & 2.22 & 0.80 & 5.00 & 21.23 & 4.64 & 20.37 & 23.41 & 0.0717 & 0.6660 \\
     BART  \cite{lewis2019bart} & 9.94 & 7.18 & 5.16 & 3.72 & 6.50 & 38.37 & 18.04 & 38.50 & 18.68 & 0.1833 & 0.8378\\
     T5 \cite{raffel2020exploring} & 42.27 & 32.00 & 24.6 & 18.58 & 29.36 & 54.16 & 31.74 & 51.24 & 43.22 & 0.2582 & 0.8841\\
     MM-MDS & 47.31 & 35.56 & 26.73 & 20.32 & 32.48 & 58.59 & 35.68 & 49.06 & 54.67 & 0.2571 & 0.9081 \\
     MM-MDS with MCS & 47.62 & 35.67 & 26.46 & 20.10 & 32.46 & 59.62 & 36.51 & 49.87 & 57.50 & 0.2644 & 0.9115\\
      MM-MDS with DI & 47.05 & 35.27 & 26.42 & 19.79 & 32.13 & 59.46 & 36.29 & 49.76 & 57.11& 0.2657 & 0.9152 \\
      MM-MDS with MCS and DI  & 47.70 & 36.48 & 27.52 & 20.85 & 33.14 & 59.74 & 37.59 & 50.93 & 58.85 & 0.2773 & 0.9144\\
    {\em MMK-Summation} with MCS & 47.57 & 35.86 & 26.98 & 20.31 & 32.68 & 59.95 & 36.49 & 50.21 & 58.03 & 0.2697 & 0.9145\\
      {\em MMK-Summation} with DI & 46.50 & 35.16 & 26.66 & 19.92 & 32.06 & 59.75 & 37.14 & 50.83 & \textbf{60.21} & 0.2746 & 0.9162 \\
    \textbf{{\em MMK-Summation}} & \textbf{48.68} & \textbf{36.85} & \textbf{27.92} & \textbf{21.50} & \textbf{33.47} & \textbf{60.86} & \textbf{37.43} & \textbf{51.05} & 58.32 & \textbf{0.2746} & \textbf{0.9180}\\
     \hline
    \end{tabular}}
    \label{R1}
    \vspace{0.5em}
\end{table*}

\begin{table*}[hbt!]
    \centering
    \caption{Performances of different baselines and proposed models for medical concern summary (MCS) generation}
    \scalebox{0.93}{
    \begin{tabular}{lccccccccccc}
    \hline
     \textbf{Model}  & \textbf{B-1} & \textbf{B-2} & \textbf{B-3} &\textbf{B-4} & \textbf{BLEU}  & \textbf{R- 1}  & \textbf{R- 2} & \textbf{ROUGE- L}  & \textbf{METEOR} & \textbf{Jaccard Sim} & \textbf{BERT Score} \\ \hline
     MM-MDS with only MCS & 29.42 & 14.97 & 1.73 & 0.16 & 11.57 & 46.11 & 20.28 & 41.1 & 30.06 & 0.1958 & 0.6365\\
      MM-MDS &  30.82 & 15.46 & 1.72 & 3.41 & 12.85 & 43.93 & 21.35 & 39.93 & 28.37 & 0.1984 & 0.5811\\
    {\em MMK-Summation} with only MCS  & 29.53 & 17.02 & 5.35 & 2.92 & 13.71 & 46.49 & 22.87 & 42.88 & 33.69 & 0.1899 & 0.6403 \\
      \textbf{{\em MMK-Summation}}  & \textbf{34.77} & \textbf{23.92} & \textbf{10.79} & \textbf{6.90} & \textbf{19.10} & \textbf{47.86} & \textbf{26.67} & \textbf{45.18} & \textbf{36.83} & \textbf{0.2295} & \textbf{0.6650} \\
     \hline
    \end{tabular}}
    \label{R2}
    \vspace{0.5em}
\end{table*}

\hspace{-0.33cm}\textbf{Key Contributions} Acquiring insights into medical concerns can facilitate doctors in formulating appropriate impressions, and the combined knowledge can contribute to crafting a comprehensive summary. Recognizing the inter-relationship among these tasks, we build a multi-tasking, multimodal medical summary generation framework that cohesively learns from all three tasks. With the immense growth of language models and the proven efficacy of huge pre-training, we leverage a pre-trained module as the backbone, incorporated with an adapter-based modality and knowledge infusion. The model receives dialogues as input, retrieves relevant external knowledge contextualized within the discourse, combines these knowledge and visual cues expressed in dialogues with the dialogue text representation, and subsequently produces MCS, DI, and overall summary. The main contributions of the work are threefold, which are enumerated below.

\begin{itemize}
    \item The work first investigates the significance of employing two-stage methods (homogeneous chunking followed by overall summarization) and multi-tasking speaker-driven approaches for medical summarization to generate a comprehensive dialogue summary.
    \item The proposed {{\em MMK-Summation}} model demonstrates superior performance compared to numerous baseline models and state-of-the-art approaches, exhibiting a substantial improvement across various evaluation metrics, including human assessment.
\end{itemize}

\section{Findings}  Based on the experiments, we report the following answers (with evidence) and observations to our investigated research questions (RQ). 

\hspace{-0.33cm}\textbf{RQ1: Is there any correlation between medical concern summary (MCS) generation and overall dialogue summarization?} To address this research question, we conducted experiments using two models: one that generates an overall summary solely based on a dialogue (MM-MDS, as shown in Table \ref{R1}), and another that simultaneously learns medical summary generation and overall summary generation tasks (MM-MDS with MCS and {\em MMK-Summation} with MCS, also shown in Table \ref{R1}). Models that jointly learn these two tasks significantly outperform the traditional model. The improvements observed across various evaluation metrics (BELU: 0.20 $\uparrow$, R1-1: 1.36 $\uparrow$, R-2: 0.81 $\uparrow$, R-L: 1.15 $\uparrow$, METEOR: 3.36 $\uparrow$, Jaccard Sim: 0.013 $\uparrow$ and BERT Score: 0.006 $\uparrow$) with {\em MMK-Summation} with MCS provide clear evidence of the benefits of enhancing overall summary generation through multitasking with medical concern summary generation. 

\hspace{-0.33cm}\textbf{RQ2: What correlation exists between doctor impression (DI) generation and overall dialogue summarization?} In clinical conversations, two primary elements stand out: patient utterances and doctor utterances. The precise summarization of patient utterances aims to capture medical concerns while summarizing doctor utterances, especially the last few, reveals the doctor's impression. Our research revealed a positive correlation between the quality of medical concern summaries and overall summarization, and vice versa. Consequently, we inferred that the doctor's impression also influences the overall summary. Interestingly, our findings indicate that the doctor's impression alone doesn't significantly contribute to the overall summary compared to using medical concern summaries alone (MM-MDS versus MMK-Summation with MCS). However, when combined with medical concern summaries, it notably enhances the performance of overall summary generation (MMK-Summation with MCS versus MMK-Summation). Upon delving into the rationale, we observed that medical concern summaries exhibit a stronger correlation with the overall summary compared to doctor impressions, evident in both word overlap and semantic similarity. Thus, based on this evidence, we assert a minor yet discernible impact of doctor impressions on overall summary generation.

\hspace{-0.33cm}\textbf{RQ3: Are these three tasks correlated? Would learning all three simultaneously result in an improved overall summary generation?} A dialogue summary is a cohesive piece of concise information encapsulating key points from different speakers engaged in the discussion. Thus, we, as humans, summarize a conversation by first summarizing each speaker's context and then forming a unified dialogue summary. Humans do it very fast, and thus, it seems like it is a one-step process. Motivated by this, we build a multitasking model, {\em MMK-Summation}, which takes dialogue as input and generates MCS, DI, and overall summary simultaneously. The proposed multitasking {\em MMK-Summation} model outperforms both MM-MDS (generating only the overall summary) and {\em MMK-Summation} with MCS/DI. The obtained improvement across different metrics are as follows: BLEU: $0.99 \uparrow$, R-1: $2.27 \uparrow$, R-2: $1.75 \uparrow$, ROUGE-L: $1.99 \uparrow$, METEOR: $3.65 \uparrow$, Jaccard Sim: $0.017 \uparrow$, and BERT Score: $0.009 \uparrow$. The improvements in terms of BLEU, ROUGE, METEOR, and BERT Score are significant for a generation task, establishing confidence in our hypothesis that learning these three tasks, MCS, DI, and overall summary generation tasks, would help in yielding an efficient overall summary.

\hspace{-0.43cm} \textbf{Key Observations} The key observations and findings are as follows: (i) The {\em MMK-Summation} model, which produces MCS, DI, and overall summaries from the decoder, surpasses the MM-MDS model with MCS/DI, which generates MCS \& DI from the encoder and overall summary from the decoder. This discovery suggests that context-aware cross-attention in the decoder notably impacts (improves) context embedding necessary for effective decoding. Additionally, it aligns with human behavior, indicating instantaneous speaker-driven chunking followed by summarization. (ii) The task of generating medical concern summaries is more closely correlated with overall summary generation than with doctor impression. (iii) We noted that conventional models like BART and T5 are notably deficient in two main areas: (a) comprehending visuals and (b) maintaining symptom-consistent diagnoses. Thus, incorporating contextualized M-modality fusion of visuals and knowledge has led to significantly superior performance in terms of all evaluation metrics, including human evaluation. 


\end{document}